\documentclass[twocolumn,letterpaper,10pt]{article}
\usepackage{iasted}
\usepackage{times}
\usepackage[dvips]{graphicx}
\usepackage[pass]{geometry}

\usepackage{amsmath}
\usepackage{amssymb}
\usepackage{graphicx}
\usepackage{multirow}
\usepackage{subfigure}
\usepackage{colortbl}
\usepackage{setspace}

\newcolumntype{V}{>{$\vcenter\bgroup\hbox\bgroup}c<{\egroup\egroup$}}

\allowdisplaybreaks[1]

\graphicspath{{./imgs/}}

\begin{document}

\date{}

\title{Computer Simulation Based Parameter Selection for Resistance Exercise}

\author{Ognjen Arandjelovi\'c\\Centre for Pattern Recognition and Data Analytics, Deakin University\\Geelong, 3216 Victoria, Australia\\\small\texttt{ognjen.arandjelovic@gmail.com}}

\maketitle
\thispagestyle{empty}

\noindent
{\bf\normalsize ABSTRACT}\newline
{In contrast to most scientific disciplines, sports science research has been characterized by comparatively little effort investment in the development of relevant phenomenological models. Scarcer yet is the application of said models in practice. We present a framework which allows resistance training practitioners to employ a recently proposed neuromuscular model in actual training program design. The first novelty concerns the monitoring aspect of coaching. A method for extracting training performance characteristics from loosely constrained video sequences, effortlessly and with minimal human input, using computer vision is described. The extracted data is subsequently used to fit the underlying neuromuscular model. This is achieved by solving an inverse dynamics problem corresponding to a particular exercise. Lastly, a computer simulation of hypothetical training bouts, using athlete-specific capability parameters, is used to predict the effected adaptation and changes in performance. The software described here allows the practitioner to manipulate hypothetical training parameters and immediately see their effect on predicted adaptation for a specific athlete. Thus, this work presents a holistic view of the monitoring-assessment-adjustment loop.}\\\vspace{-5pt}

\noindent
{\bf\normalsize KEYWORDS}\newline weight, training, muscle, strength, powerlifting

\vspace{-5pt}\section{Introduction}\vspace{-2pt}
\label{s:intro} Sports science is a discipline characterized by a strong focus on practical application. Ultimately, the aim of any research in this field is to facilitate advancement in some aspect of the athletic endeavour. The nature of such advancement may take on many forms. An improvement in performance may be achieved through the use of a novel training modality \cite{SwinLloyAgouStew2009} or better training parameter selection \cite{RobbYounBehmPayn2010}, for example. Alternatively, strategies to enhance intra-training \cite{CaruCoda2008} or inter-training \cite{SharPear2010} recovery rates may be devised. Injury prevention methods \cite{ColaGarc2009} or methods for accelerating rehabilitation \cite{FithPoweKhan2010}, over time albeit indirectly can also be seen to contribute to improved performance. While certainly not an exhaustive list, the
aforementioned elements of an integral training regime have been attracting the most attention from researchers and practitioners. The complexity emerging from the interrelatedness of these elements illustrates the breadth of potential avenues for further study and potential scientific contribution to the sports community.

In broad terms, the development of a novel idea in sport science comprises three distinct challenges before reaching the stage of general acceptance by the practitioners. The first of these concerns the pursuit of data collection by means of empirical study. Indeed, this aspect of research has been dominating sports science for most of its existence, producing a consistently expanding corpus of available data. The accumulation of empirical findings facilitates the second challenge -- the understanding of the underlying physiological mechanisms. This is achieved by the unification of regularities in the observed data by means of phenomenological models. Such models effectively reduce the total information content needed to describe a particular phenomenon and are subjected to scrutiny through the predictions they produce. In this final stage the model is applied in practice i.e.\ athletic training.

This paper focuses on the final of the aforementioned developmental stages. Specifically, it considers several outstanding problems associated with the application of a recently proposed physiological model underlying resistance training performance and adaptation. These involve the estimation of measurable performance characteristics from realistic and only loosely constrained videos of athletes in training, the process of estimation of free model parameters from said characteristics, and the subsequent use of the model to guide future training choices in a manner tuned to a specific athlete.

\vspace{-7pt}\section{Performance extraction from real video}\vspace{-2pt}\label{s:characteristics}
The central concept in the computational model introduced in \cite{Aran2010-med} is the capability profile of an athlete in a given exercise. It is instrumental in predicting performance as well as in capturing the nature and magnitude of training adaptations. An athlete's capability profile $\hat{F}$ for a given exercise is
defined as the maximal force that the athlete can exert against the load in the exercise as a function of the load's position (commonly elevation) $\delta$ and velocity $v$:
{\small\begin{align}
  \hat{F} \equiv \hat{F}(\delta,v).
\end{align}}
It can be thought of as a generalization of the force-length and force-velocity characteristics of an isolated skeletal muscle to an arbitrary exercise \cite{Aran2010-med}. Force-length and force-velocity characteristics, while trainable \cite{CaioPerrEdge1981} and variable between different people as well as across different muscles of the same person, generally share the same functional form. However, this is not the case for a capability profile corresponding to an arbitrary exercise. The universal characteristics of force production for individual muscles are modulated by the plurality of the involved musculature, attachment structure of individual muscles and the change in the biomechanics throughout the lift.

\begin{figure*}
  \vspace{-10pt}
  \centering
  \subfigure[]{\includegraphics[height=0.2\textwidth]{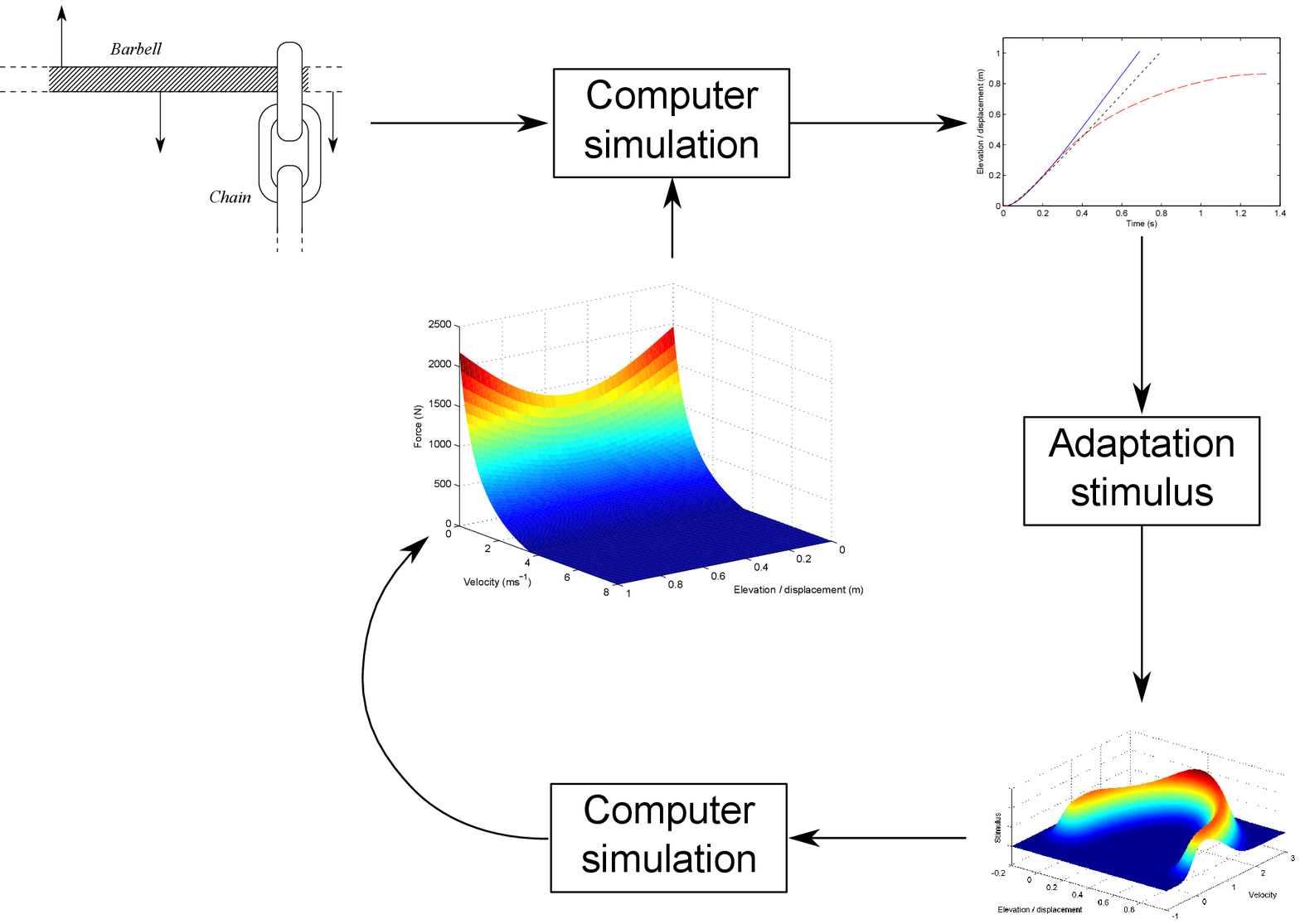}\label{f:modeloverview}}
  \subfigure[]{\includegraphics[height=0.2\textwidth]{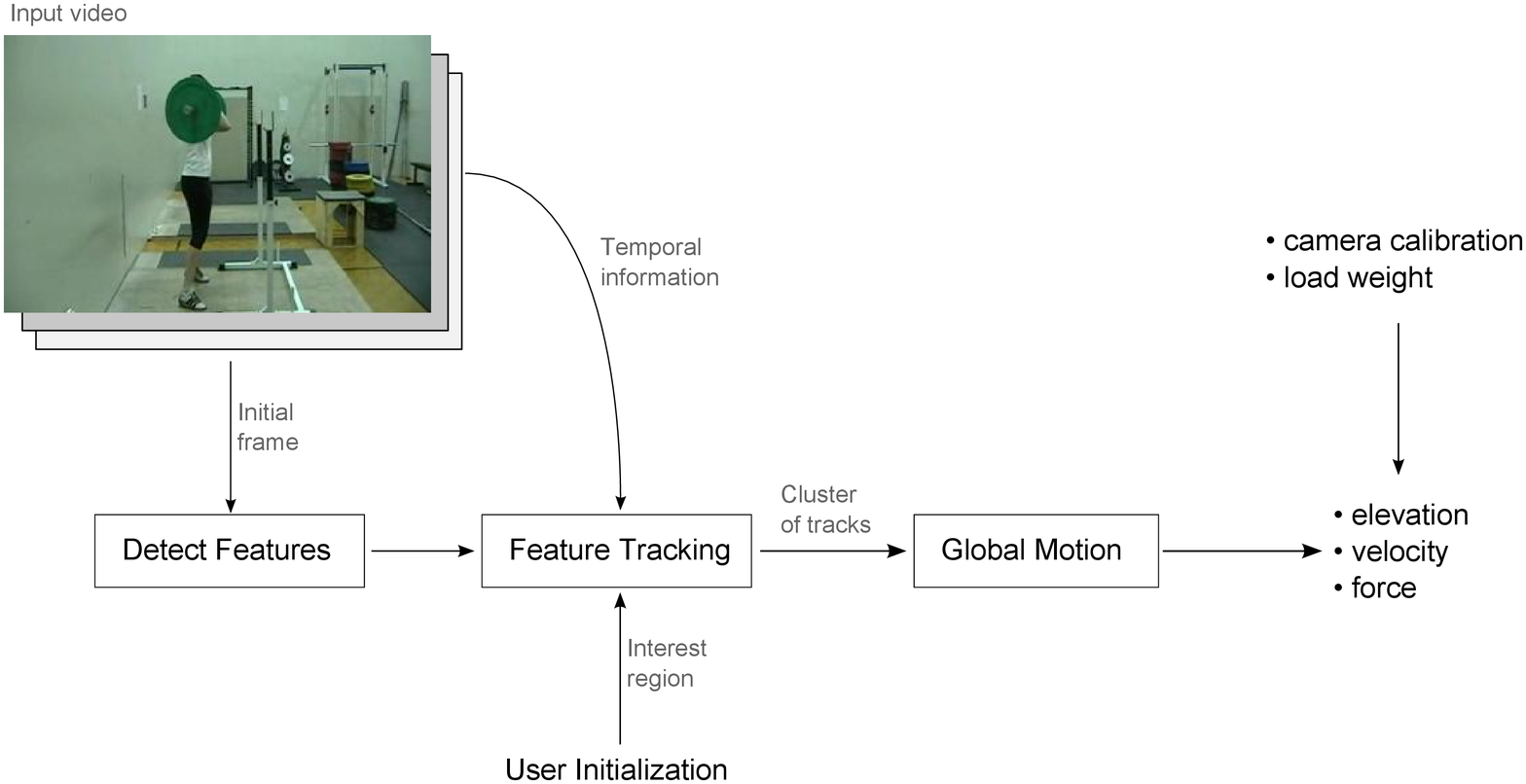}\label{f:overview}}
  \subfigure[]{\includegraphics[height=0.2\textwidth]{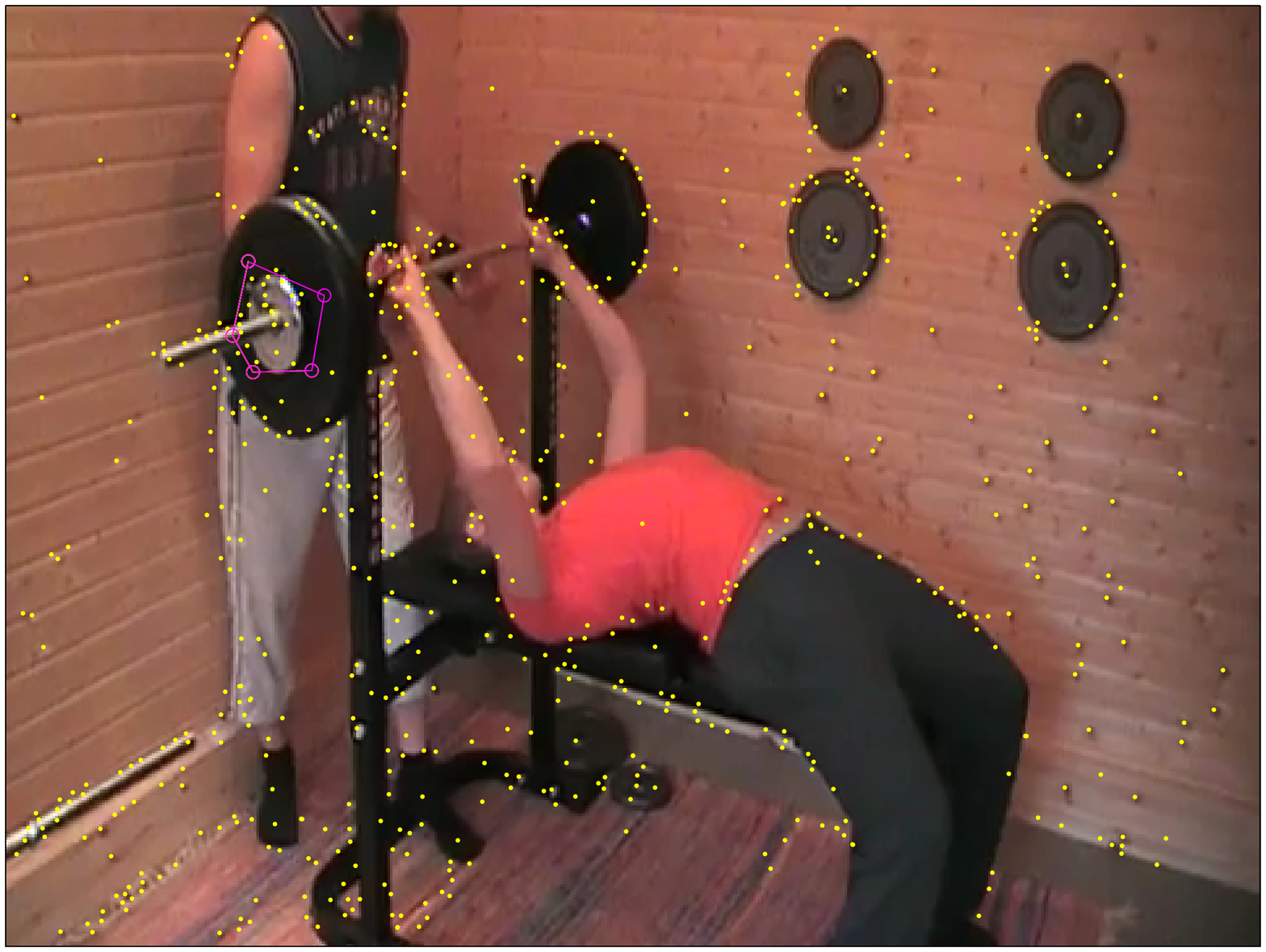}\label{f:input}}
  \vspace{-5pt}
  \caption{\footnotesize (a) Overview of the computational model adopted from \cite{Aran2010-med}. (b) Key elements of the proposed motion extraction. (c) Original frame with detected features (yellow dots), and the interest region outlined by the user (purple line).}
  \vspace{-20pt}
\end{figure*}

The model is employed by predicting exercise performance first. Using a numerical approximation to the differential equation governing the motion of the load, a computer simulation is applied to predict the motion of the load through time. The force exerted on the load during the
movement is explicitly given by the athlete's capability profile, exponentially modulated by the accumulated fatigue. Simulation results are then used to infer the adaptational stimulus, which manifests itself through a fed-back modification of the capability profile. The entire training-adaptation loop is
summarized in Fig.~\ref{f:modeloverview}.

Herein the aim to show how this model can be utilized in practice. The specific challenges addressed are (i) the estimation of the free model parameters from data which can be acquired without specialized equipment, large cost or effort, and (ii) the application of the model in the context of a computer program for practical training planning.

\vspace{-5pt}\subsection{Approach Overview}\vspace{-2pt}
Our first contribution is an algorithm for estimating the motion of the load in weight-lifting exercise. A summary of the key components of the proposed method is shown in Fig.~\ref{f:overview}. First, \emph{interest points} in the starting frame of the input video are detected. Overlaid on the original image these are displayed to the user who selects a region corresponding to the load used for exercise. Then, each interest point within this region of interest is tracked until the completion of the video producing a series of continuous \emph{motion tracks}, one for each interest point. Information from all extracted tracks is polled together to infer reliably the overall motion of the load which is then processed further to extract the effective force exerted against the load. The starting and terminal times of individual repetitions are extracted here too.

\vspace{-5pt}\subsection{Interest point detection}\vspace{-2pt}\label{ss:pts}
Let $\mathbf{F}_1$ be the greyscale image representation of the initial frame of an input video sequence and {\small$\mathbf{F}(x,y)$} (or equivalently {\small$\mathbf{F}(\mathbf{x})$}) to the intensity of the pixel at the image location {\small$(x,y)$}. Then the corresponding \emph{Gaussian scale-space} {\small$\mathbf{S}(x,y,s)$} is a three-dimensional volume defined as:
{\small\begin{align}
  \mathbf{S}(x,y,s) = \mathbf{F}_1(x,y) \ast_{xy} \mathbf{G}(x,y;s\sigma)
\end{align}}
where $\ast_{xy}$ denotes convolution over $x$ and $y$, and $G(x,y;s\sigma)$ is an isotropic 2D Gaussian, with the covariance $(s\sigma)^2$. The scale parameter $s$ governs the degree of image blur and suppresses image features of lesser spatial extent than $\approx \sqrt{s}$. From the scale-space, interest points are detected at the loci of maximal rate of appearance change with scale which are also spatially well localized. This means that they are local extrema across space and scale of the difference of Gaussian-smoothed images. This initial list is further narrowed down by accepting only those loci which are well localized by requiring both eigenvalues of the corresponding Hessian at the detection scale to be sufficiently large \cite{Lowe2004}. Low-contrast loci or line-like regions are thus filtered out. A typical result is shown in Fig.~\ref{f:input}.

\vspace{-5pt}\subsection{Feature seeding}\vspace{-2pt}
By construction, interest points are image loci with locally characteristic appearance. As such, they are promising candidates for reliable tracking of motion through time. However, our specific aim here is to extract the motion of the load lifted by the athlete -- the video sequence may contain other, confounding sources of motion which are not of interest (e.g.\ other trainees). Thus, we seek to restrict our attention to those interest points which are within the region corresponding to the moving load.

The initialization of the tracking is difficult to automate fully because the load can greatly vary in appearance: it may comprise a fixed dumbbell or a loaded barbell, while the plates used to load it can differ in their shape, dimensions and colour. Thus, we adopt a semi-automatic approach, whereby brief user-input is used to initialize the tracker: the initial frame of the video sequence is displayed and the user asked to outline a region of the image corresponding to the load used by the athlete. The loci of detected interest points, which are marked on the displayed image, thus serve to guide the user who can choose a region with their maximal number, as in Fig.~\ref{f:input}

\vspace{-5pt}\subsection{Feature cluster tracking}\vspace{-2pt}
Having located a set of discriminative image loci of interest, the goal is to track them over time. The methodology employed here is similar to \cite{TomaKana1991}. There are two key differences in the approach taken here: in the initialization of the tracked windows and in the search for the optimal frame-to-frame wrapping parameters. Unlike in \cite{TomaKana1991} where the choice of tracking windows is based on the spatio-temporal gradient matrix corresponding to the first two video frames, here the tracked regions surrounding interest points are detected as described in Sec.~\ref{ss:pts}. The size of each square region is set equal to the detection scale of its interest point.

As in the previous work, tracking is formulated as an optimization problem, whereby the region of interest $\mathcal{W}$ in frame $\mathbf{F}_i$ is localized in the subsequent frame $\mathbf{F}_{i+1}$ by estimating the set of parameters $\mathbf{a}\in \mathbb{R}^6$ of an affine transformation which maps $\mathcal{W}$ onto a region in $\mathbf{F}_{i+1}$, such that the observed image difference is minimized. A modification introduced here is to estimate $\mathbf{a}$ using a three-level pyramidal
coarse-to-fine scheme whereby the initial estimate is made using quarter-resolution images, which is then refined at half-resolution and finally full resolution. This serves both to increase the speed of convergence as well as the robustness of the estimate by preventing the iterative gradient descent (described next) from getting stuck to a locally optimal value. Formally, at each level of the pyramid, we wish to minimize:
{\small\begin{align}
  e(\mathbf{a}) = \sum_{\mathbf{x} \in \mathcal{W}}  \big[~\mathbf{F}_{i+1}(\mathbf{x}_a) - \mathbf{F}_i(\mathbf{x})~\big]^2 ,
  \label{e:KLerror}
\end{align}}
where $\mathbf{x} = [x~y]^T$ is an image locus, and:
\renewcommand{\arraystretch}{0.7}
{\small\begin{align}
\mathbf{x}_a = \left[
                     \begin{array}{c}
                       (1+a_1)~x +     a_3~y + a_5 \\
                           a_2~x + (1+a_4)~y + a_6 \\
                     \end{array}
                   \right].
  \label{e:affine_warp}
\end{align}}
Minimization of the error term $e(\mathbf{a})$ is a non-linear optimization task which can be solved through an iterative steepest descent scheme. Using the first order Taylor series expansion of the expression in Eq.~\eqref{e:KLerror} results in a quadratic minimization problem that can be solved in closed form. A simple analysis shows that minimal $\hat{e}(\mathbf{a})$ is achieved for:
{\small\begin{align}
   \Delta \mathbf{a} =
   &\left\{ \sum_x \left(\frac {\partial \mathbf{x}_a} {\partial \mathbf{a}}\right)^T \nabla {\mathbf{F}_{i+1}}^T
    \nabla \mathbf{F}_{i+1} \frac {\partial \mathbf{x}_a} {\partial \mathbf{a}}  \right\}^{-1} \\
    &\sum_x \left[\nabla \mathbf{F}_{i+1} \frac {\partial \mathbf{x}_a} {\partial \mathbf{a}} \right]^T
    \bigg[\mathbf{F}_i(\mathbf{x}) - \mathbf{F}_{i+1}(\mathbf{x}_a) \bigg] \notag
   \label{e:KLdeltap}
\end{align}}
Eq.~\eqref{e:KLdeltap} and the warping parameter update are applied until convergence i.e.\ until the magnitude of the update fails to exceed an error tolerance threshold {\small$\|\Delta \mathbf{a}_i\| \leq \epsilon$}.

\vspace{-5pt}\subsection{Robust motion estimation}\vspace{-2pt}
The algorithm described in the previous section tracks a particular local feature within the region of interest, initially corresponding to an automatically detected interest point. However, generally, the region of interest contains many features, each of which produces a track {\small$\mathcal{T}_n = x_n(0), \ldots, x_n(k_n \Delta t)$} (for {\small$n$-th} feature). As indicated by different maximal time step indices {\small$k_i$}, the tracks may be of different durations -- a feature once lost in tracking is not re-spawned.

\begin{figure}[thp]
  \vspace{-10pt}
  \centering
  \footnotesize
  \subfigure[]{\includegraphics[width=0.22\textwidth]{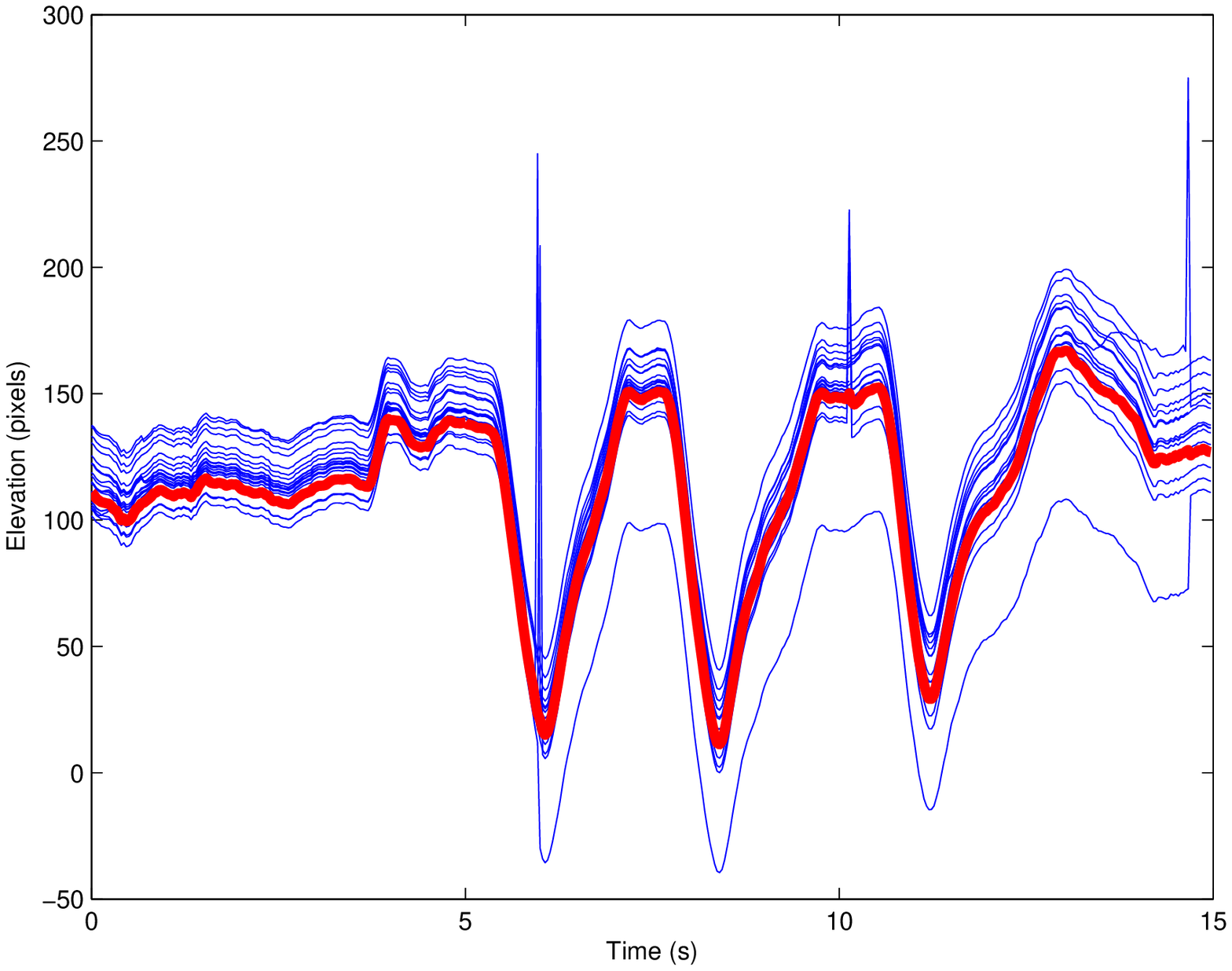}}~~~~
  \subfigure[]{\includegraphics[width=0.22\textwidth]{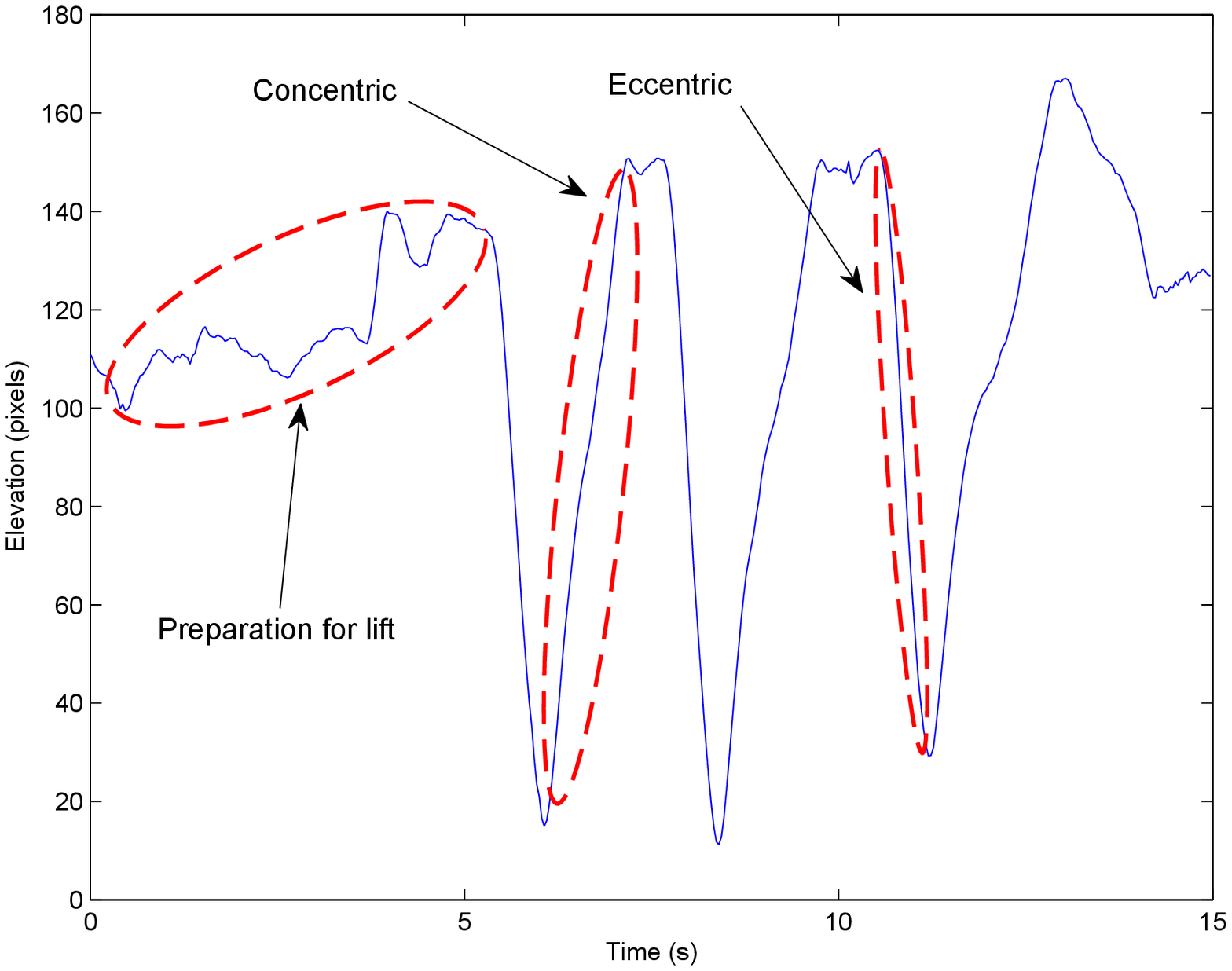}}
  \subfigure[]{\includegraphics[width=0.22\textwidth]{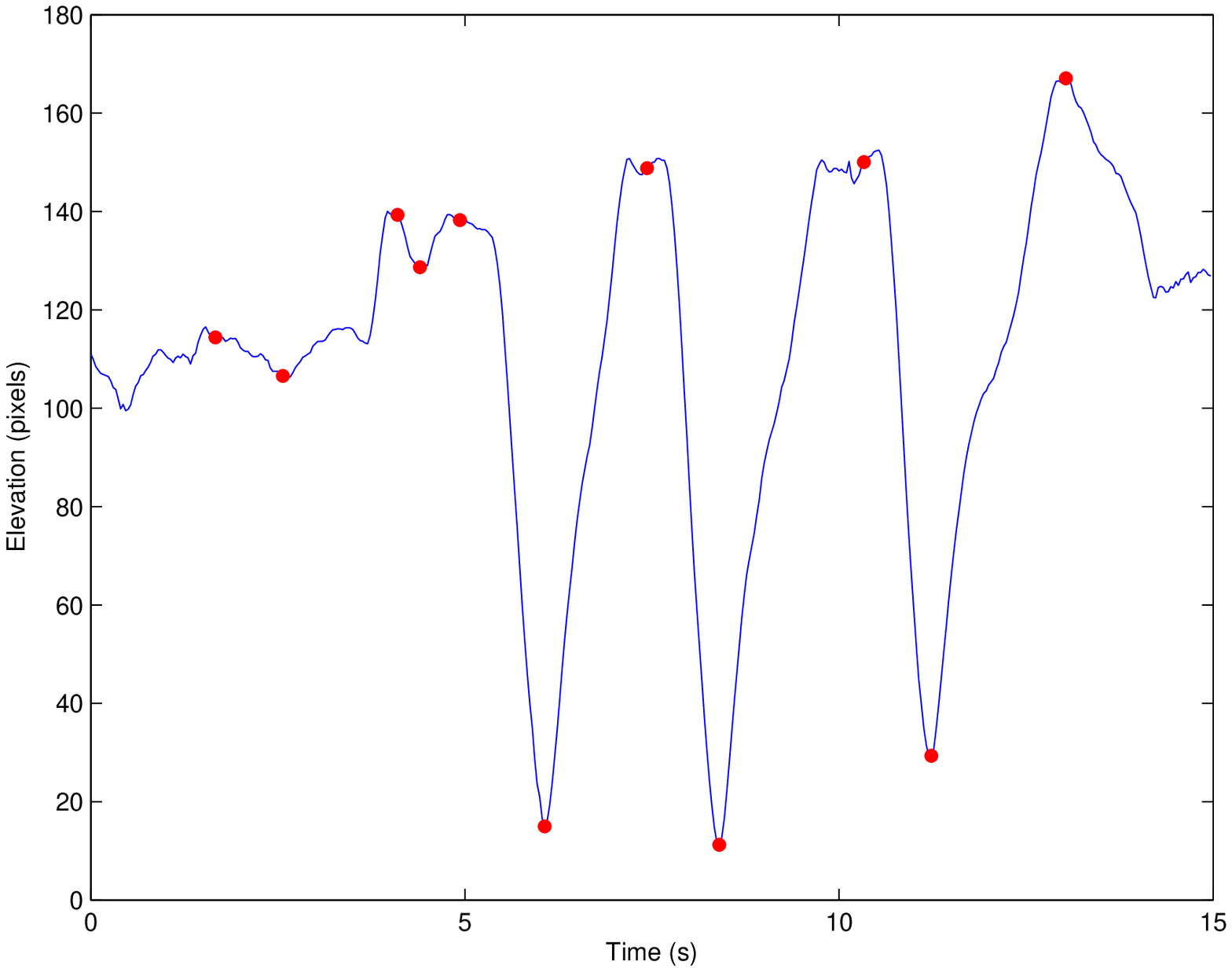}}~~~~
  \subfigure[]{\includegraphics[width=0.22\textwidth]{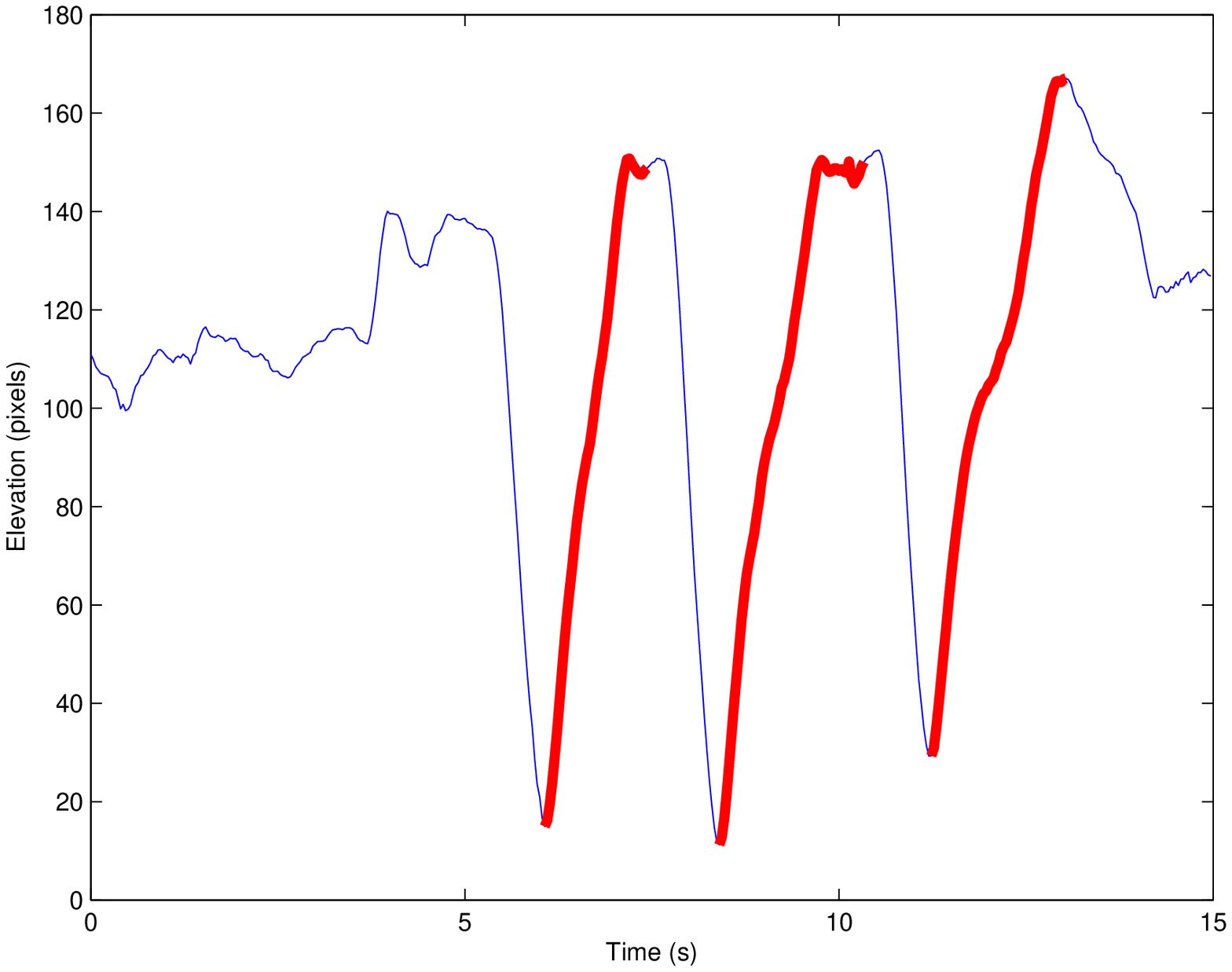}}
  \vspace{-5pt}
  \caption{\footnotesize (a) Tracks of vertical displacement over time of features of interest (thin blue
            lines) and the estimated robust overall motion (thick, red line).
            (b) Overall vertical displacement track, marked with semantic labels of different stages of exercise. (c) The detected starting and terminal points of the concentric portions of each completed repetition (red circles and dotted vertical lines). (d) Variation in the vertical displacement of the load during three extracted concentric bouts.}
  \label{f:motion}
  \vspace{-20pt}
\end{figure}

An example of a set of tracks extracted from a typical lifting video sequence is shown in Fig.~\ref{f:motion}(a). Each thin line is the vertical track of a single feature. Note different starting values of elevation of different features' tracks -- these correspond to different initial locations and are not of relevance here. It is the coherence in their relative motion which is being exploited in computing the mean load displacement, shown as the superimposed thick red line.

\begin{figure}[thp]
  \centering
  \footnotesize
  \subfigure[]{\includegraphics[height=0.15\textwidth]{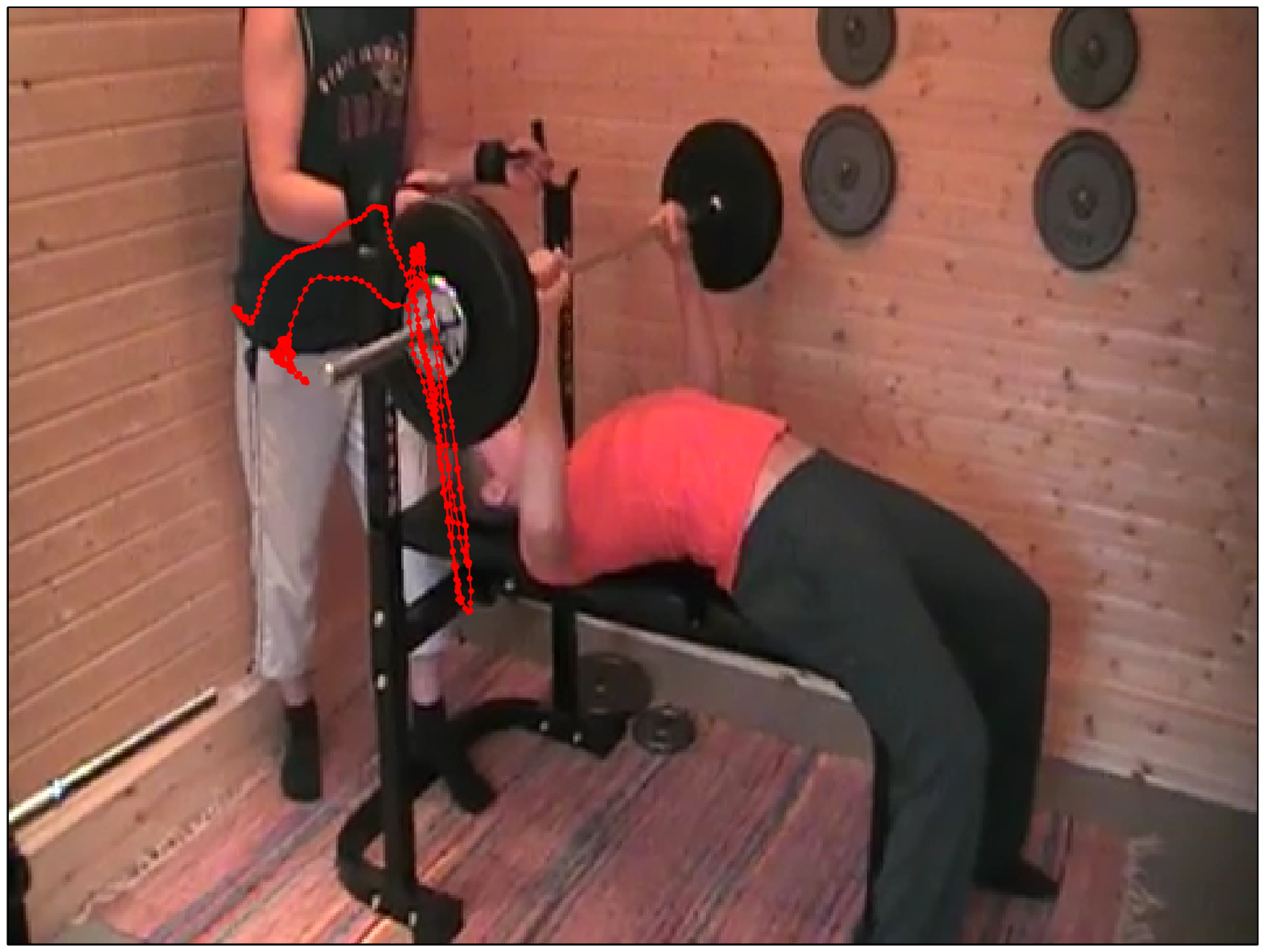}}
  \subfigure[]{\includegraphics[height=0.15\textwidth]{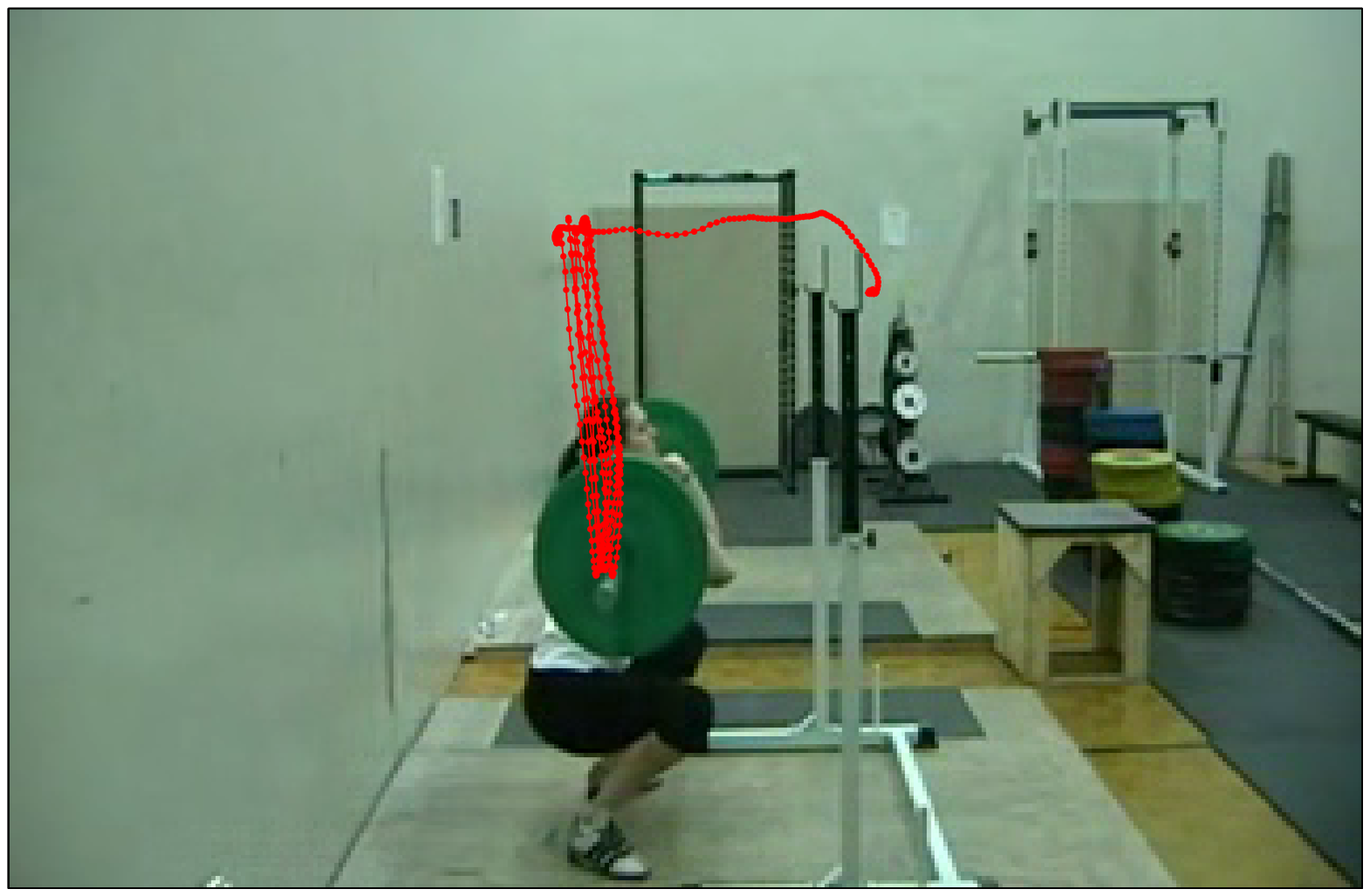}}
  \caption{\footnotesize Examples of load tracking -- the extracted motion of the load is overlaid as a red line on a typical video frame.}
  \label{f:examples}
  \vspace{-20pt}
\end{figure}

As will become apparent in Sec.~\ref{s:capability}, precise tracking of the load is crucial for the accurate estimation of the variation in the force exerted by the trainee. Here we use the entire set of obtained feature tracks to infer more robustly their shared translatory motion, that is, the motion of the rigid load they correspond to. Let, without loss of generality, {\small$k_1 \leq k_2 \leq \ldots \leq k_n$}. We compute the location of the load at time {\small$k\Delta t$} as follows. If {\small$D_k$} are displacements at {\small$k\Delta t$} at most {\small$\Delta$} pixels from the median displacement:
{\small\begin{align}
  V_k &= \left \{~ x_i(k \Delta t) - x_i(0) ~:~ k_i \leq k ~ \right\}\\
  D_k &= \left \{~ d ~:~ d \in V \wedge \|d - \mu_{1/2}(V)\| \leq \Delta ~ \right\},
\end{align}}
the load displacement is computed as the robust mean:
{\small\begin{align}
  \bar{d}(k \Delta t) = \frac{1}{|D_k|}\sum_{d \in D_k} d
\end{align}}
A typical result is illustrated in Fig.~\ref{f:motion} and Fig.~\ref{f:examples}.

\vspace{-5pt}\subsubsection{From image displacements to physical motion}\vspace{-2pt}
Hitherto we only concerned ourselves with the image motion of the load. As our final goal is to model quantities which exist in the physical world, such as the force producing capability of an athlete, we need to link the apparent motion {\small$x(t)$} with actual physical motion {\small$\delta(t)$}. In general, this is an ill-posed problem -- the process of imaging, that is to say of projecting 3-dimensional geometry of the physical world onto a 2-dimensional image plane, inherently creates ambiguity. This ambiguity can be resolved only by imposing further constraints, specific to a particular task. Specifically, in this work we consider lifts in which the only relevant resistive forces are constrained to the vertical direction (note that this does not mean that the motion of the load is constrained to the vertical direction). Most obviously this applies to free weight lifts, which are resisted by the force of gravity, but can also include a variety of other machine based exercises with frictional, elastic and viscous resistive forces (see Sec.~\ref{s:application}). Consequently, since all that is needed for the estimation of velocities, acceleration and forces involved is relative motion, i.e.\ displacement, assuming that the extent of any horizontal motion of the load is small compared to the load's distance from the camera, the relationship between the two quantities {\small$x(t)$} and {\small$\delta(t)$} is:
{\small\begin{align}
  \delta(t) = K_x x(t)
\end{align}}
The value of the multiplicative constant {\small$K_x$} is determined through simple calibration using a
known reference object (e.g.\ the length of a standard Olympic barbell).

\vspace{-5pt}\subsection{Concentric motion extraction}\vspace{-2pt}
The extracted motion of the load includes different aspects of a lifting bout, as illustrated in Fig.~\ref{f:motion}(b). Initially, the athlete is preparing for the lift and the load may exhibit motion as the athlete assumes a comfortable starting pose. This is then followed by alternating eccentric and concentric lifting efforts (not necessarily in that order) separated by usually brief pauses i.e.\ isometric holds. They facilitate the dissipation of some of the
accumulated fatigue, allow the athlete to focus on the forthcoming repetition, catch breath, check positioning etc. Static holds may follow either the eccentric or the concentric portion of the lift, depending on the exercise biomechanics.


\vspace{-5pt}\section{Capability profile reconstruction}\vspace{-2pt}\label{s:capability}
In the previous section we saw how the variation in the elevation of the load used for resistance exercise can be robustly extracted from video without strong assumptions on the exercise, viewpoint or the appearance of the load. Here our goal is to use these measured performance characteristics to infer the athlete's exercise-specific fitness, that is, in the context of the performance model adopted in this paper, the athlete's \emph{capability profile}.

\vspace{-5pt}\subsection{Estimating velocity and force variation}\vspace{-2pt}\label{ss:estVelForce}
The athlete's capability profile $\hat{F}$ in an exercise is defined as a bivariate function capturing the dependence of the maximal force that the athlete can exert against the load and the load's elevation $\delta$ and velocity $v$. That is:
{\small\begin{align}
  \hat{F} \equiv \hat{F}(\delta,v).
\end{align}}
We wish to infer this variation from a set of motion tracks, each corresponding to a concentric portion of a repetition in a given lift, extracted using the algorithm from Sec.~\ref{s:characteristics}.

Consider the vector comprising the displacement (elevation) and velocity of the load over time, $\mathbf{\delta}(t) \equiv \left[ \delta(t)~~\dot{\delta}(t)\right]^T$ where a dot over a symbol signifies time differentiation (thus $\dot{\delta} = \frac{d\delta}{dt}$ is the rate of change of elevation, or velocity, and $\ddot{\delta} = \frac{d^2 \delta}{dt^2}$ is the rate of change of velocity, or acceleration). This \emph{state vector} of the load changes throughout the lift, thus making a path $\mathcal{P}$ through the two-dimensional elevation-velocity (or capability) plane. The idea proposed here is that the capability profile $\hat{F}(\delta,v)$ can be inferred in the localities of all available paths $\mathcal{P}_i$ from the estimates of the effective force variation $F_i(t)$ along the said paths.

\vspace{-5pt}\subsection{Velocity, acceleration and effective force}\vspace{-2pt}\label{ss:vaF}
The quantity directly measured from video is the load elevation. From the position of the load, its vertical velocity must be estimated to obtain capability plane tracks $\mathcal{P}_i$, as well as its acceleration from which $F_i(t)$ can be computed.

In principle, this involves simple differentiation of the variation in the elevation $\delta(t)$. Given samples from $\delta(t)$ at discrete and equidistant intervals $t \equiv t_k = k\Delta t$, the velocity can be estimated using the standard three-point finite difference approximation to a derivative:
{\small\begin{align}
  &\hspace{-120pt}v_{k} =
  \begin{cases}
     \frac{1} {2\Delta t} \cdot (\delta_{k+1} - \delta_{k-1})& :~ k > 0\\
     0                                                       & :~ k = 0 ~~~~~\text{ (initial condition)}
  \end{cases}\\
\text{and similarly the acceleration:} \notag\\
  &\hspace{-120pt}a_{k} =
  \begin{cases}
     \frac{1} {2\Delta t} \cdot (v_{k+1} - v_{k-1})& :~ k > 0\\
     \frac{1} {\Delta t}  \cdot (v_{k+1} - v_{k})&   :~ k = 0\\
  \end{cases}
\end{align}}
where the subscript $k$ is used to denote the value of a particular variable at time $t = k\Delta t$. However, this approach has the undesirable effect of amplifying high frequency noise present in the initial estimates of $\delta_k$ \cite{AlonCuadLugrPint2010}. The corruption of the desired signal is particularly pronounced with
repeated differentiation. On the other hand, the usual practice of simple data smoothing prior to differentiation is problematic because it can result in physiologically unrealistic force estimates \cite{SilvAmbr2002}. Instead, to ensure that our known physical constraints are satisfied, we fit a constrained smoothing cubic spline to load elevation values $\delta(k\Delta t)$ and then differentiate the spline itself. Specifically, we construct a spline which minimizes the objective function which comprises two terms: (i) the discrepancy between the observed data (load elevation) and that predicted by the spline, and (ii) the spline roughness. Formally, the objective function $\epsilon_{\delta}$ is:
{\small\begin{align}
  \epsilon_{\delta} = \omega \underbrace{ \sum_{k} \left| \delta(k\Delta t) - \hat{\delta}(k\Delta t) \right|^2}_{\text{Fitting disagreement}} + (1-\omega) \underbrace{ \int_t \ddot{\delta}(t)^2 dt }_{\text{Roughness}},
\end{align}}
and the initial condition constraint $\dot{\delta}(t)=0$.

\vspace{-5pt}\subsubsection{Estimation of effective force}\vspace{-2pt}
The final step in the process of extracting lift characteristics from video proposed in this paper is the estimation of effective force exerted by the athlete and against the load. Having estimated the variation of the load's position, velocity and acceleration through time, force can be computed from the differential equations
of motion, that is, by the method of so-called ``inverse dynamics''. In its most general form, the motion of the load can be described through an equation capturing the dependency of its position $\delta$ on (i) the force $F$ applied against the load, (ii) the velocity, acceleration and possibly higher order derivatives of the load's position, and (iii) a set of exercise parameters $\mathbf{\Lambda}$ which include a variety of biomechanical variables. Formally:
{\small\begin{align}
  0 = \psi(F, \delta, \dot{\delta}, \ddot{\delta}, \ldots~;\mathbf{\Lambda})
  \label{e:invdyn}
\end{align}}
The application of inverse dynamics then comprises the computation of $F$ from the known values of the remaining quantities in Eq.~\eqref{e:invdyn}.

\vspace{-5pt}\subsection{Fatigue modelling and parameter inference}\vspace{-2pt}\label{ss:maxExertionModel}
In the previous section it was shown how the effective force $F(t)$ exerted by the lifter can be estimated from the motion of the load and the prior knowledge of the system dynamics. Under the adopted model, this force is bounded above by the value of the capability profile for the corresponding state $\hat{F}(\delta(t),\dot{\delta}(t))$, modulated by the accumulated fatigue:
{\small\begin{align}
  F(t) \leq \hat{F}(\delta(t),\dot{\delta}(t)) \cdot \exp(-t/T_F),
  \label{e:effectiveF}
\end{align}}
where $T_F$ is the person-specific fatigue time constant, unknown \textit{a priori}. In simulations reported in \cite{Aran2010-med}, and the discussion of a possible approach for model parameter inference, it was assumed that the upper bound in Eq.~\ref{e:effectiveF} was actually attained at all times. In other words, the athlete was assumed to always attempt to maximally accelerate the load. This assumption was justified by the focus of the original publication on strength and power athletes, such as powerlifters, who indeed do observe this practice in training \cite{BehmSale1993}. However, the aim in the present work is to devise an approach more widely applicable and, as will be shown, the aforementioned assumption of continuous maximal exertion does not hold well for maximal sets at intensities lower than $\approx 85\%$.

\vspace{-5pt}\subsubsection{Variable fatigue model}\vspace{-2pt}
Firstly, to account for non-maximal exertion, Eq.~\eqref{e:effectiveF} is here extended to explicitly account for a variable rate of fatigue accumulation. Formally:
{\small\begin{align}
  F(t) = \hat{F}(\delta(t),\dot{\delta}(t)) \cdot g(t),
\end{align}}
where $0 \leq g(t) \leq 1$ is the newly introduced fatigue modulating function, and:
{\small\begin{align}
  \frac {d g(t)} {dt} = - \frac{1} {T_F} ~ g(t) ~ \rho(t)
\end{align}}
The coefficient $\rho(t)$, where $0 \leq \rho(t) \leq 1$, effectively scales the time fatigue constant from its minimal value of $T_F$ attained during maximal exertion. The rate of maximal voluntary force loss is decreased at the time of submaximal effort (lower force can be sustained longer):
{\small\begin{align}
  \rho(t) = \hat{F}(\delta(t),\dot{\delta}(t)) ~/~ F(t).
\end{align}}
Note that when $\rho(t) \equiv 1$ i.e.\ when $\hat{F}(\delta(t),\dot{\delta}(t)) \equiv F(t)$, the form of the fatigue function becomes simply $g(t) = \exp(-t/T_F)$, as in the original model \cite{Aran2010-med}.

\vspace{-5pt}\subsubsection{Force-fatigue management model}\vspace{-2pt}\label{sss:fatigueManagement}
In \cite{Aran2010-med}, it was assumed that in training the athlete whose performance and adaptational responses were modelled, at each point in the lift exerts the maximal force possible. This force is readily computed using the athlete's capability profile corresponding to the lift in question and the model of fatigue accumulation. The assumption of continuous maximal exertion is effectively a simple model of force-fatigue management, characterizing how an athlete employs the underlying capability to produce force to complete the lift. In this work, an alternative model is described which is aimed at a broader range of athletes. Our focus is on athletes who explicitly seek performance improvement across a range of intensities, unlike powerlifters who are ultimately concerned only with performance at the maximal intensity i.e.\ 100\% of 1RM.

Here we consider \emph{trained athletes}. This allows us to assume that the use of the underlying force production capability is approximately optimized for the training task. Specifically, we assume that for a given training intensity (i.e.\ load relative to 1RM), the athlete's force production is such as to complete a
repetition with \emph{minimize fatigue accumulation} thus allowing the athlete to perform the most work (repetitions) at this intensity.

\begin{figure}
  \centering
  \includegraphics[width=0.30\textwidth]{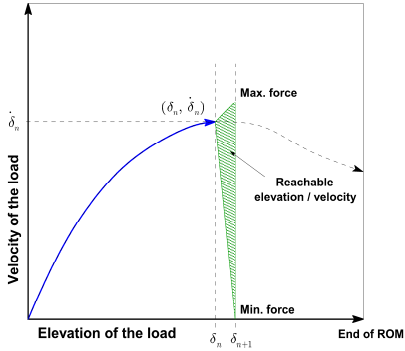}
  \caption{\footnotesize  At every point in a successful lift the motion of the load is constrained. This
            can be visualized usefully by considering the lift as a path through the capability plane (see Sec.~\ref{s:characteristics}). In the process of inference of the underlying capability profile, the position of the load at $(\delta_n, \hat{\delta}_n)$ constrains its position at the next time increment to a triangle, defined by the load's current position $(\delta_n, \hat{\delta}_n)$, the point which would be reached in the case of subsequent immediate failure $(\delta_{n+1}, 0)$ and the point which would be reached in the case of athlete's maximal exertion. }
  \label{f:optimization}
  \vspace{-10pt}
\end{figure}

To formalize the above, let $L^{(k)}(\delta_{n+1},v)$ be the negative logarithm of the fatigue modulating function $g(t)$ at the repetition $k$, the load's position $\delta_{n+1}$ and velocity $v$:
{\small\begin{align}
  L^{(k)}(&\delta_{n+1},v) = - \log g(t).
\end{align}}
Then, to meet the assumption of the minimal accumulated fatigue, the force exerted by the athlete at each time step $n$ has to satisfy:
{\small\begin{align}
  L^{(k)}(&\delta_{n+1},v) = \min_{v' \in \mathcal{R}_s} \left[~ L(\delta_n,v') + \frac{\Delta t}{T_F} \cdot \rho(t) ~\right] \\
  = &\min_{v' \in \mathcal{R}_s} \left[~ L(\delta_n,v') + \frac{\delta_{n+1}-\delta_n}{T_F (v + v')/2} \cdot
            \frac{F_n(\delta_n,v)}{\hat{F}(\delta_n,v) }
            ~\right] \label{e:athletemodel}. 
\end{align}}
Here, fatigue corresponding to $L^{(k)}(\delta_{n+1},v)$ is minimized by considering the minimal fatigue achievable at the previous time step at $L^{(k)}(\delta_n,v')$ and the incremental increase in fatigue accumulated in reaching $L^{(k)}(\delta_{n+1},v)$ from $L^{(k)}(\delta_n,v')$.  The range of possible velocities $v'$ at the previous time step is restricted by the athlete's ability to produce force to a region $\mathcal{R}_s$ in the capability plane. This concept is graphically illustrated in Fig.~\ref{f:optimization} which shows the locus $(\delta,\dot{\delta})$ in the capability plane, the path though the capability plane corresponding to the preceding stages of the repetition (blue arrow), and the region of interest for the next time step (shaded, green). This region is triangular and defined by the locus $(\delta_n,\dot{\delta}_n)$, the condition that failure does not take place (i.e.\ the locus $(\delta_n,0)$) and the maximal velocity that the load can have given the athlete's capability (corresponding to the maximal force that the athlete can produce). Finally, the repetition has to end with the load velocity $v_T$ such that:
{\small\begin{align}
  v_T = \arg \min_{v_T} L^{(k)}(\delta_{\text{max}},v_T).
\end{align}}
This boundary condition enforces global optimality of the repetition i.e.\ minimizes the total accumulated fatigue.

\vspace{-12pt}\paragraph*{Inference}
Observe that the nature of lifting performance optimality described by Eq.~\eqref{e:athletemodel} is not such that incremental fatigue at each time step is minimized, that is, the term $\frac{\delta_{n+1}-\delta_n}{T_F (v + v')/2} \cdot \frac{F_n(\delta_n,v)}{\hat{F}(\delta_n,v) }$. Rather than being local, optimization is \emph{global}. It is by virtue of this assumption that it is necessary to constrain our attention to trained individuals \cite{ShimKraeSpieVole2006}. Specifically, the reader should note that fatigue minimization described by the introduced model is not achieved through conscious efforts of the athlete. Instead, it is an adaptation of the neuromuscular system induced through repeated training bouts.

In mathematical vernacular, the optimization problem of interest is not ``greedy''. On the other hand, it does exhibit the property of \emph{optimality of nested overlapping subproblems}. This is readily apparent by inspection from Eq.~\ref{e:athletemodel} -- the optimal solution at the load position $\delta_{n+1}$ can be expressed as a function of a locally computable term and the optimal solution corresponding to the position of the load at the preceding time step, that is, $\delta_n$. Optimization problems of this type are solvable efficiently. However, note that this is not what what we are trying to achieve here. Rather than trying to compute the optimal solution, our goal is to infer the underlying model parameters (the athlete's capability profile) from (i) the optimal solution and (ii) the form of the model. The optimal solution is given by the lifting characteristics or, equivalently, the corresponding repetition paths in the capability plane. The form of the underlying model is that described by Eq.~\ref{e:athletemodel}. The difficulty of this inference is rooted in the global nature of the optimization, that is, in terms of our mathematical model, the loss of local information through summation.

\vspace{-12pt}\paragraph*{Failure}
Consider the last attempted repetition in a set which ends in momentary muscular failure. Referring back to the illustration in Fig.~\ref{f:optimization}, in the last elementary time interval $\Delta t$, the shaded region $\mathcal{R}_s$ collapses to a line -- the velocity of the load drops to $0$ even when maximal possible force is applied by the athlete. This means that the coefficient $\rho(t)$ is equal to $1$. By means of mathematical induction and working backwards in time, it can be seen that $\rho(t)=1$ for the entire duration of the final repetition. Thus, we can write:
{\small\begin{align}
  F_n(\delta_n,v) &= \hat{F}(\delta_n,v) \cdot \exp\left\{ -L^{(K)}(\delta_n,v) \right\}\notag \\
  &=
  \hat{F}(\delta_n,v) \cdot \exp\left\{ -L^{(K)}(0,0) - t/T_F
  \right\},\label{e:lastrep}
\end{align}}
where $K$ is the index of the final repetition. It is clear from Eq.~\ref{e:lastrep} that the values of the capability profile $\hat{F}(\delta,v)$ along the path corresponding to the final repetition can be computed directly up to scale.

\vspace{-12pt}\paragraph*{Successful repetitions}
The lifting conditions during the last repetition in a set are rather special -- failure to complete the lift results despite athlete's maximal effort investment. In contrast, the preceding, successful repetitions offer a ``choice'' (not necessarily conscious, as noted earlier) in the manner force exerted against the load is managed over time, This choice is described mathematically in the form of the optimization in Eq.~\ref{e:athletemodel}. It is the global nature of this optimization which makes lifting characteristics measured during successful repetitions less informative in the reconstruction of the underlying capability profile. Successful repetitions can merely be used to formulate a lower bound on the values of the capability profile along the capability plane paths corresponding to the repetitions. For this
reason, in the present work, successful repetitions are not used in the capability profile reconstruction.

\vspace{-12pt}\paragraph*{Multiple sets}
Repetitions of sets performed by the same athlete but different intensities trace different paths in the capability plane. Thus, performance characteristics at a range of training intensities can be used to infer the functional forms of different regions of the athlete's capability profile underlying the exercise in consideration.

In practice, sufficient data for accurate reconstruction of the region of the capability profile relevant to the athlete's performance could be accumulated with ease. This is especially true in the case of cornerstone exercises (e.g.\ bench press, squat) which are practiced with relatively high frequency and volume. Monitoring training performance over only a few sessions would typically suffice. In this paper, to overcome the limited amount of data we had available and extend the area of the capability plane over which capability is estimated, we also employ interpolative and extrapolative methods. These are employed while ensuring the conformance of the results with constraints derived to the fundamental physiological principles underlying the capability profile. Specifically, we require that the capability profile is monotonically decreasing in the ``velocity direction'' i.e.\ that for any given point in an exercise, maximal effective force that the athlete can exert against the load decreases with the increase in its velocity:
{\small\begin{align}
  \forall \delta, v_1 < v_2~:~ F(\delta,v_1) > F(\delta,v_2)
  \label{e:monoton}
\end{align}}
For a single muscle, Eq.~\eqref{e:monoton} follows trivially from Hill's equation. For an arbitrary number of contributing muscles in a complex, compound exercise, the same conclusion follows from Hill's equation and the monotonicity of functions $\Psi$ and $\Phi_i$ which are in \cite{Aran2010-med} used to model exercise biomechanics and impose kinematic constraints.

\begin{figure}[thp]
  \centering
  \includegraphics[width=0.35\textwidth]{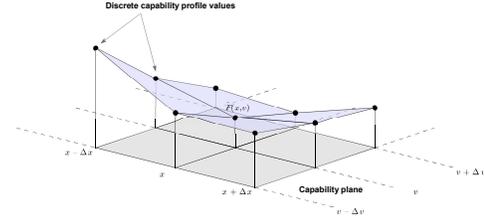}
  \caption{\footnotesize  The capability profile is represented by a set of samples taken from a regular dense mesh over the capability plane. }
  \label{f:cpsamples}
  \vspace{-25pt}
\end{figure}

Recall from Sec.~\ref{ss:vaF} and \cite{Aran2010-med} that a capability profile is represented by a set of samples. As illustrated in Fig.~\ref{f:cpsamples}, the samples correspond to predetermined, discrete values of the load's position and velocity i.e.\ a regular, dense mesh over the capability plane. As explained earlier in this section, only those samples which lie on the paths of set-ending repetitions are directly measured. To estimate the values of the capability profile corresponding to regions enclosed by the paths, interpolation using a quadratic form penalty was performed. Formally, the discrepancy in the values of the capability profile of two samples neighbouring in the $x$ or position direction is computed as:
{\small\begin{align}
  \Delta J_\delta = k_\delta \left[ \hat{F}(\delta,v) - \hat{F}(\delta + \Delta \delta,v) \right]^2.
\end{align}}
Similarly, for samples neighbouring in velocity and diagonal directions:
{\small\begin{align}
  &\Delta J_v = k_v \left[ \hat{F}(\delta,v) - \hat{F}(\delta,v + \Delta v) \right]^2\\
  &\Delta J_{\delta v} = k_{\delta v }\left[ \hat{F}(\delta,v) - \hat{F}(\delta + \Delta \delta,v + \Delta v) \right]^2.
\end{align}}
Thus, the full error function $J$ which is minimized is:
{\small\begin{align}
  J =
          &\sum_{\delta=0}^{\delta_{\text{max}}- \Delta \delta} \sum_{v=0}^{v_{\text{max}}} k_\delta \left[ \hat{F}(\delta,v) - \hat{F}(\delta + \Delta \delta,v \right]^2 \\
        &+ \sum_{\delta=0}^{\delta_{\text{max}}} \sum_{v=0}^{v_{\text{max}}- \Delta v} k_v \left[ \hat{F}(\delta,v) - \hat{F}(\delta,v+ \Delta v) \right]^2~\notag \\
        &+ \sum_{\delta=0}^{\delta_{\text{max}}- \Delta \delta} \sum_{v=0}^{v_{\text{max}}- \Delta v} k_{\delta v} \left[ \hat{F}(\delta,v) - \hat{F}(\delta+ \Delta \delta,v+ \Delta v) \right]^2.\notag
\end{align}}
As the form of $J$ is quadratic, minimization over unknown values of the capability profile samples is computed readily in closed form by differentiation.

\vspace{-5pt}\subsection{Cross-validation and empirical results}\vspace{-2pt}
In this paper we introduced a cascade of algorithms which allow for an athlete's capability profile to be estimated from the athlete's resistance training performance captured in video form. Our methods need only minimal human input and allow for the use of realistic and virtually unconstrained video sequences. Thus, little technical proficiency from the user is required. We finish this section with an empirical demonstration of how the underlying capability profile representation together with the algorithms developed in the present work allows for accurate and principled prediction of performance under unseen conditions.

\vspace{-5pt}\subsubsection{Capability profile estimation}\vspace{-2pt}\label{sss:cpestimate}
In the case of all video sequences used for the evaluation herein, the camera angle was not in any way specially chosen (e.g.\ to capture either the fully frontal or the fully profile view of the trainee). As desirable in practice, the camera was instead simply placed in a location which was found to be convenient in the context of the equipment setup of the training facility.

\begin{figure}[thp]
  \centering
  \subfigure[]{\includegraphics[width=0.22\textwidth]{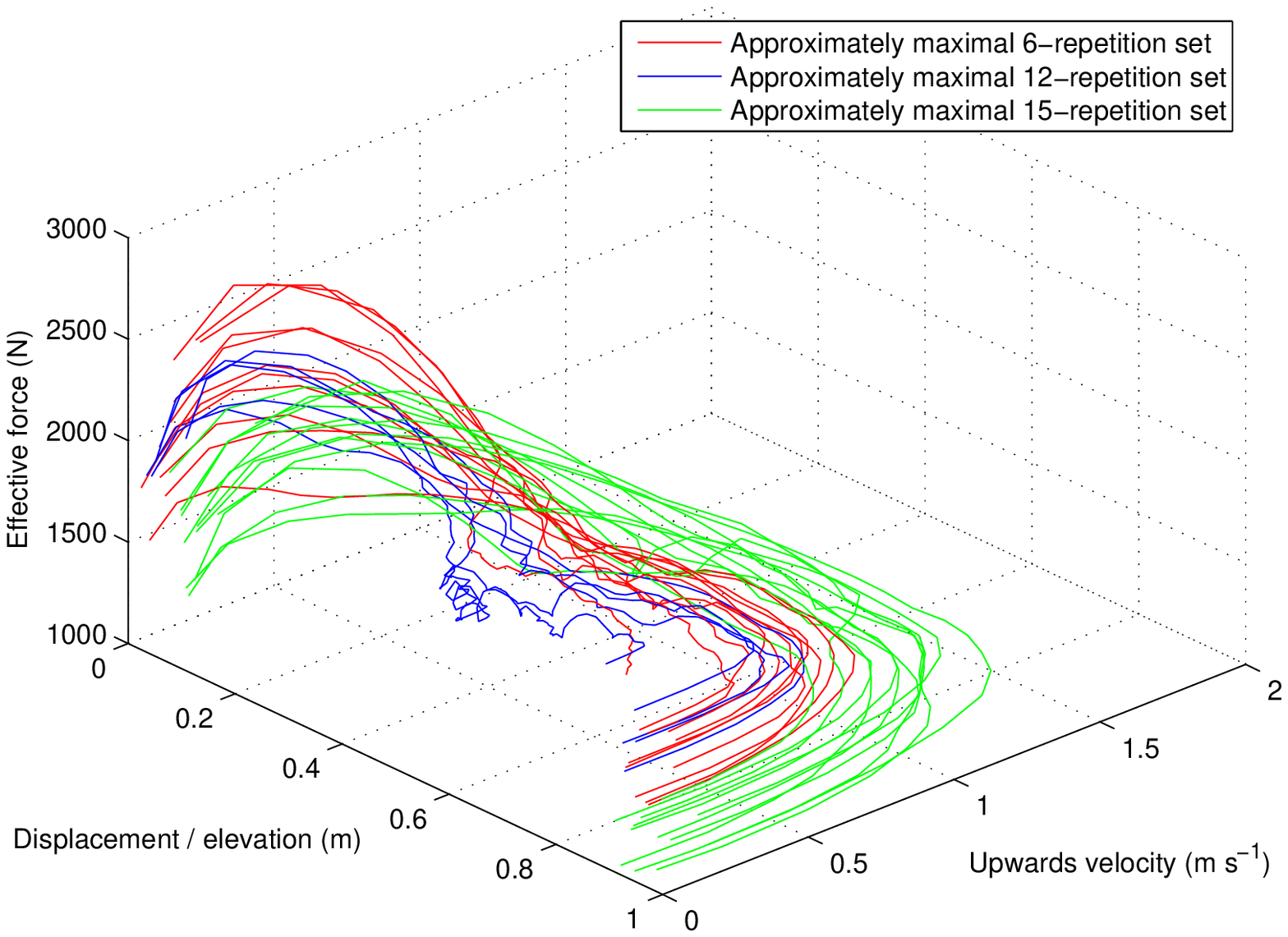}\label{f:paths3D}}
  \subfigure[]{\includegraphics[width=0.22\textwidth]{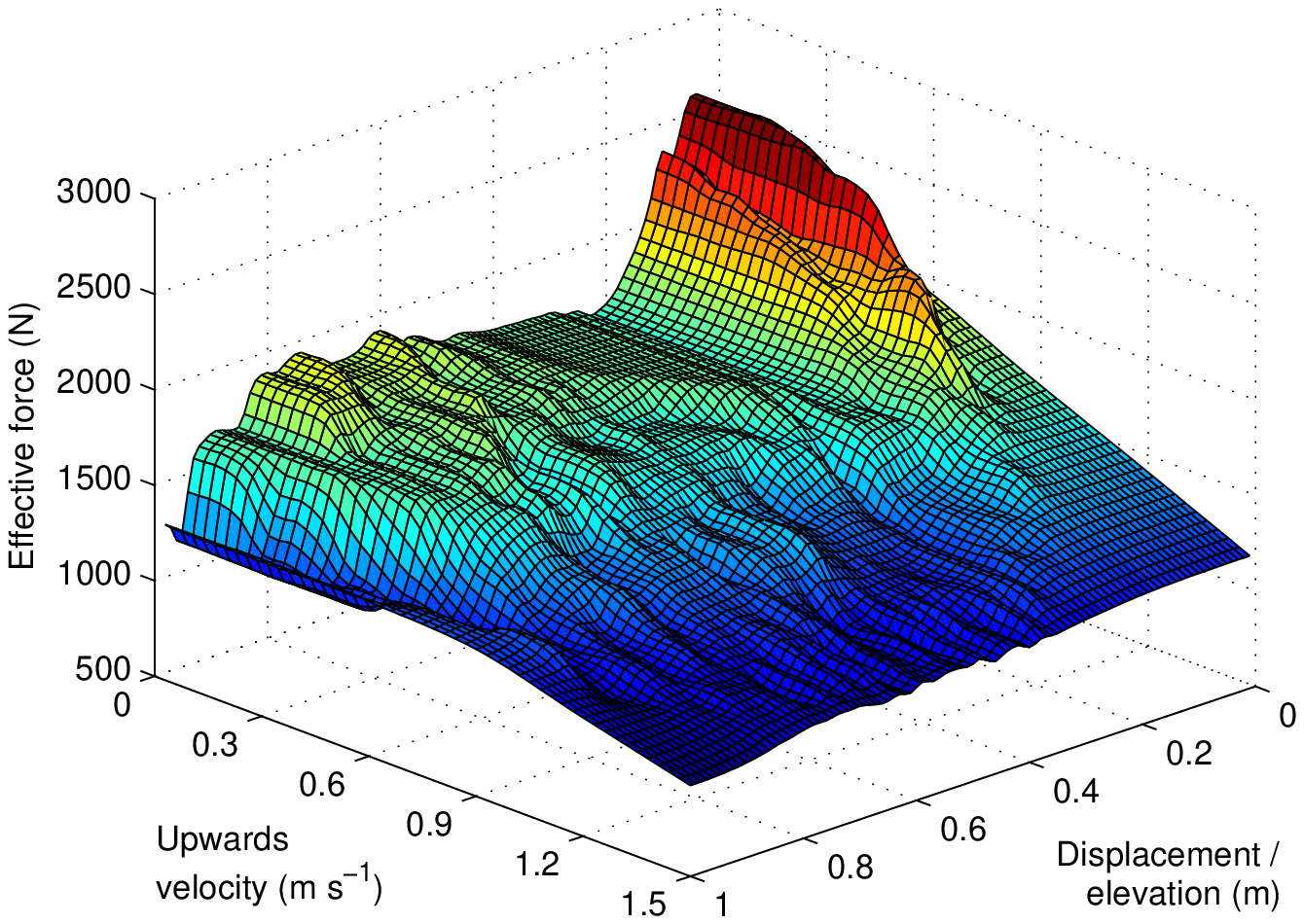}\label{f:reconstruction}}
  \caption{\footnotesize (a) Variation in effective force exerted by the athlete against the load through repetitions of sets at different loading intensities and (b)
  the corresponding capability profile reconstruction. }
  \vspace{-20pt}
\end{figure}

An example of the extracted training lift characteristics is shown in Fig.~\ref{f:paths3D}. Note that the remarkable resemblance of the characteristics of different repetitions in the same set supports our fatigue management model. Under the maximal exertion model used in \cite{Aran2010-med}, greater effects of fatigue would have been expected. The capability profile reconstruction is shown in Fig.~\ref{f:reconstruction}.

\vspace{-5pt}\subsubsection{Comparison with Brzycki equation predictions}\vspace{-2pt}
The Brzycki equation \cite{Brzy1998} is a well known equation which predicts the maximal number of repetitions an athlete can complete with a particular load based on a maximal effort test at a different intensity. One of its most common applications of the Brzycki equation is to estimate an athlete's 1RM load (i.e.\ $100$\% intensity) since maximal strength is a useful and readily understood performance indicator relevant in many athletic disciplines \cite{WhisPantEastBroe2003}. The Brzycki equation captures a simple regression model. If $w_{n_{\text{rep}}}$ is the measured maximal $n_{\text{rep}}$-repetition load, this model predicts the absolute maximal load (1RM)  $\hat{w}_1$ as:
{\small\begin{align}
  \hat{w}_1 = 36~\frac{w_{n_{\text{rep}}}} { 37 - n_{\text{rep}} } \approx
  \frac{w_{n_{\text{rep}}}} { 1.0278 - 0.0278 ~ n_{\text{rep}} }.
  \label{e:brzycki}
\end{align}}
Applying the prediction using the measured $n_{\text{rep}} = 12$ repetition maximum of $w_{n_{\text{rep}}} = 275.0$~lbs gives the estimated 1RM of $396.0$~lbs. This estimate was compared with that of the proposed model. Single repetition lifting efforts were simulated iteratively, with progressively increasing loads until the lowest load at which failure occurs was reached ($\pm 0.5$~lbs). The 1RM of the trainee established by means of the described simulation was found to be $391.0$~lbs which is in close agreement with Brzycki's prediction. This is particulary impressive considering that empirically obtained maximal strength estimates for high repetition ranges (such as the 12RM) exhibit greater test-retest variability \cite{StocBeckDefrDill2011}.

\vspace{-5pt}\subsubsection{Comparison with measured performance}\vspace{-2pt}\label{sss:compWithReal}
While a comparison of maximal effort lifting performances predicted using statistical, regression techniques and that using the model proposed in the present paper allows for clear and readily understood validation of the information extracted as a capability profile, the capability profile model is much richer in information, allowing for a far wider spectrum of predictions to be cast. To exemplify this, here we also show an example of a comparison between the actual, empirically measured performance characteristics with those simulated using the capability profile estimate of Sec.~\ref{sss:cpestimate}.

\begin{figure}[thp]
  \centering
  \vspace{-10pt}
  \includegraphics[width=0.3\textwidth]{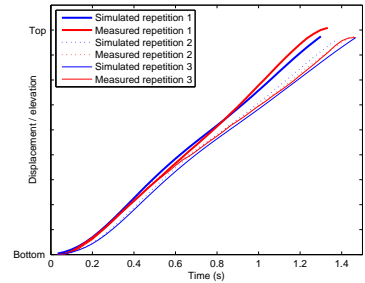}\label{f:comparison}
  \caption{\footnotesize Predicted (blue) and actual (red) 3RM performance to exhaustion, shown as plots of the load's elevation against time. The resulted prediction exhibits remarkable agreement with measured performance at both low frequencies (i.e.\ long time scale) and high frequencies (i.e.\ as subtle motion features at short time scales). }
  \vspace{-20pt}
\end{figure}

Actual lifting performance characteristics were collected by asking the trainee to perform the maximal number of repetitions using a 3RM load which was previously determined to be $375$~lbs. From a video recording of the lift, the elevation and velocity of the load through time were then extracted using the methods described in Sec.~\ref{ss:estVelForce}. Finally, a comparison was made with performance simulated using the capability profile of Sec.~\ref{sss:cpestimate}. The result of this comparison is summarized on the graph shown in Fig.~\ref{f:comparison}. Lifting characteristics predicted by the model described in this paper match the measured motion of the load remarkably well throughout the entire duration of the lift i.e.\ across all three repetitions. It is particularly interesting to observe that the model correctly predicted even subtle phenomena such as small convexities and concavities in the elevation-time plots of Fig.~\ref{f:comparison}. The convexities and concavities likely correspond to loci in the exercise ROM when a transition, respectively, from a biomechanically weaker to a biomechanically stronger or a biomechanically stronger to a biomechanically weaker position of the load occurs. That performance characteristics of this nature are predicted with such precision provides strong evidence that the underlying model is capable of accurately capturing those elements of the athlete's fitness which govern relevant exercise performance, as well as that the proposed methodology for inferring the parameters of the model is extracting meaningful information from training data.

\vspace{-5pt}\section{Application in training analysis and design}\vspace{-2pt}\label{s:application}
Owing to the central role that the capability profile plays, the
ability to estimate it from actual performance opens a wide range of
possibilities for practical use. To illustrate this, we developed a
computer application that allows a practitioner to investigate
predicted athlete-specific effects of differently targeted training
regimes. The key aspects of the application's functionality are
described next.

\vspace{-5pt}\subsection{Summary of software features}\vspace{-2pt}
Fig.~\ref{f:appMod}(a) shows the main window of the software and its principal elements. The window consists of four panels and a selection of buttons controlling the application. The panel furthest to the left is the Capability Profile Panel which displays the capability profile which is studied. Furthest to the right is the Exercise Setup Panel containing controls that adjust a variety of exercise parameters (that are not already implicitly incorporated in the capability profile). The central two panels display simulated performance characteristics (as in Sec.~\ref{sss:compWithReal}), predicted by using the capability profile shown in the Capability
Profile Panel and resistance variables from the Exercise Setup Panel. The first of the central panels shows predicted performance as a plot of the load's elevation against time; the other panel shows the same data but in the form of the corresponding capability plane path (the reader may find it useful to revisit the material of
Sec.~\ref{ss:estVelForce} as well as \cite{Aran2010-med}).

\vspace{-5pt}\subsubsection{Capability profile}
\begin{figure}[thp]
  \centering
  \subfigure[]{\includegraphics[width=0.42\textwidth]{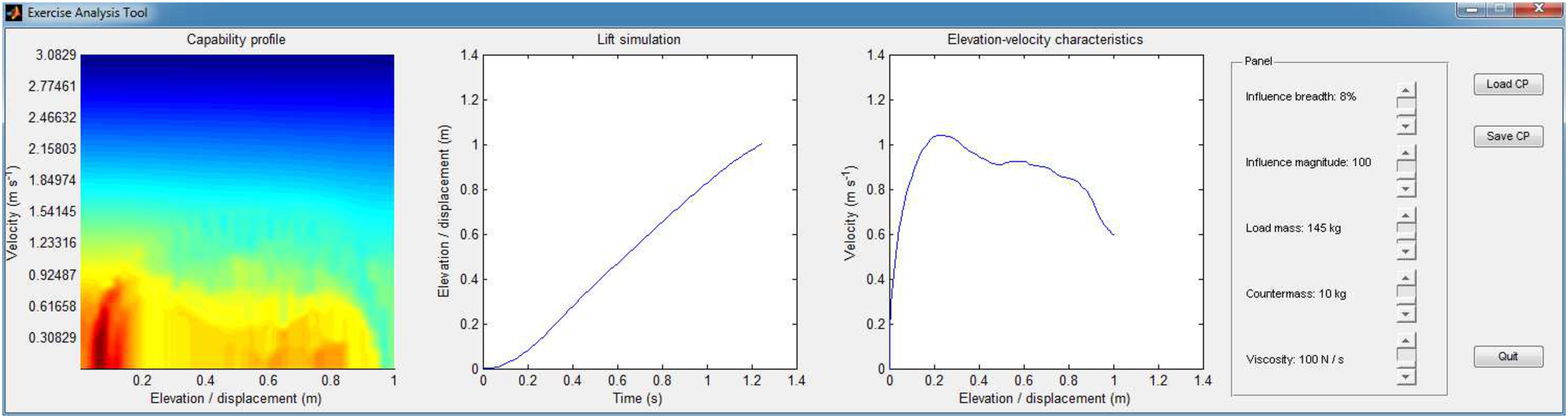}}
  \subfigure[]{\includegraphics[width=0.42\textwidth]{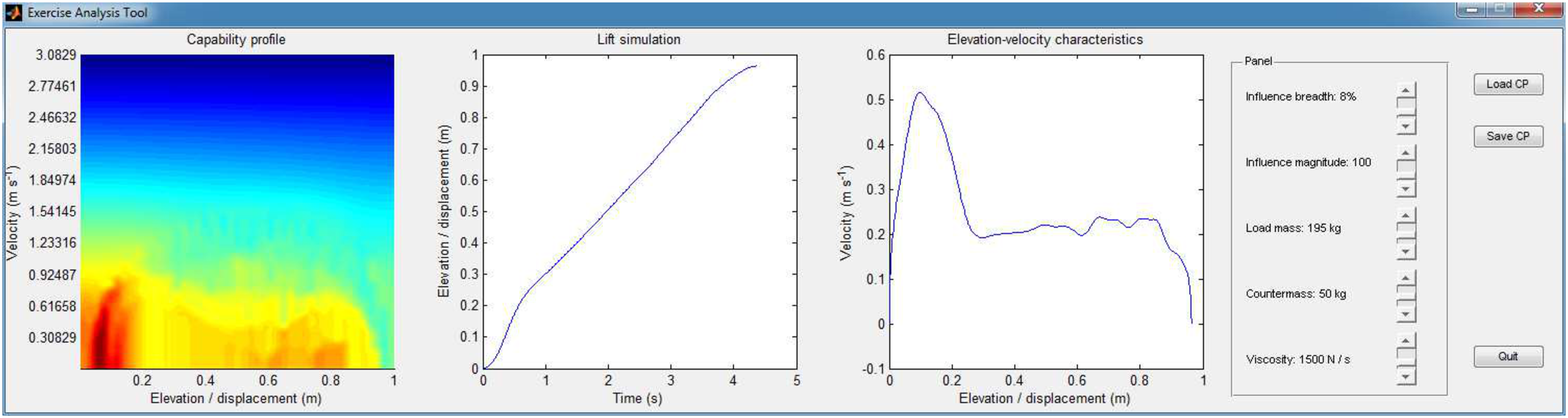}}
  \subfigure[]{\includegraphics[width=0.42\textwidth]{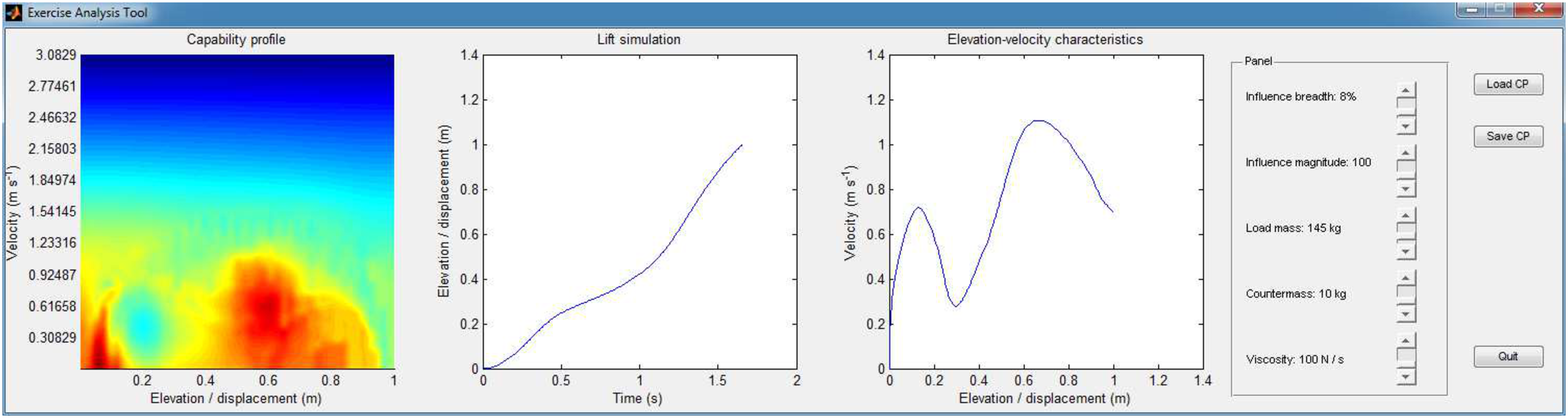}}\vspace{-5pt}
  \caption{\footnotesize Software demonstrating a practical application the proposed methods. (a) Capability profile displayed as a colour-coded image and the predicted performance characteristics for the values of resistance parameters in the right hand side panel. Performance characteristics are automatically reevaluated and visually updated when (b) the resistance settings are changed or when (c) the user modifies the capability profile. }
  \label{f:appMod}
  \vspace{-10pt}
\end{figure}

\vspace{-2pt}
The left-most panel in the main window of our software application shows the capability profile, displayed as an image. The rate of force production at a particular combination of values of the load's elevation and velocity is indicated using a colour-code, with warmer colours corresponding to higher force and cooler colours to lower force, see Fig.~\ref{f:appMod}(a). Note, for example, that the top of the image is uniformly blue corresponding to diminished capability to exert force against a rapidly moving load.

The capability profile, which may have been estimated using the algorithm described in the previous sections, can be modified by the user. Clicks in the capability plane with the left and right mouse buttons produce, respectively, positive and negative Gaussian ``bumps'' in the profile. Formally, a click at the location corresponding to $(x_0,v_0)$ creates a modified profile $\hat{F}_{\text{mod}}(\delta,v)$ from $\hat{F}(\delta,v)$:
{\small\begin{align}
  \hat{F}_{\text{mod}}(\delta,v) = \hat{F}(\delta,v) + \nu \cdot G(\delta_0,v_0; \sigma_\delta, \sigma_v),
  \label{e:adj}
\end{align}}
where $\nu$ is the adjustable magnitude of the effect, while parameters $\sigma_\delta$ and $\sigma_v$, which too are user-adjustable, control its breadth in the capability plane. This principle of capability profile modification is similar to that in \cite{Aran2011-med}.

The example in Fig.~\ref{f:appMod}(c) shows the resulting capability profile after the original one from Fig.~\ref{f:appMod}(a) was modified by decrementing the force in the locality of the point of elevation-velocity $(\delta,v) \approx (0.2~\text{m},0.45~\text{m~s}^{-1})$ and incrementing it in the locality of $(\delta,v) \approx (0.6~\text{m},0.6~\text{m~s}^{-1})$.

\vspace{-5pt}\subsubsection{Exercise setup}
\begin{figure}[thp]
  \centering
  \scriptsize
  \includegraphics[width=0.30\textwidth]{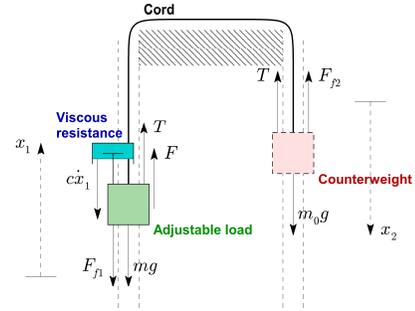}
  \caption{\footnotesize Schematic diagram of the key components of the resistance mechanism considered herein. Forces acting on the system when the direction of the velocity of both the adjustable load and the counterweight is positive (and thus, respectively, in the upward and downward directions, due to differently oriented axes measuring the two displacements). }
  \label{f:sm}
  \vspace{-20pt}
\end{figure}

\vspace{-2pt}
The right-most panel of the main window is the Exercise Setup Panel, used to control a number of exercise parameters. The first two of these control the effects of user input. ``Influence breadth'', changes the width of the capability profile modification affected by input:
{\small\begin{align}
  \sigma_\delta = \text{(infl.\ breadth)} \times \delta_{\text{max}} &&
  \sigma_v = \text{(infl.\ breadth)} \times v_{\text{max}}
\end{align}}
``Influence magnitude'' controls the magnitude $\nu$ of the adjustment in Eq.~\eqref{e:adj}. The remaining parameters control the nature of resistance used to predict performance characteristics achieved when the current capability profile is used in a computer simulation of a lifting effort. To account for different types of resistance commonly found in weight training equipment we consider the general mechanism schematically illustrated in Fig.~\ref{f:sm}. ``Load mass`` is the mass $m$ of the free adjustable load, e.g.\ a barbell, while ``countermass'' $m_0$ and ``viscosity'' $c$ are respectively the mass of a counterweight and the viscous resistance constant. A free weight lift is obtained for $m_0 = 0$ and $c = 0$.

\vspace{-5pt}\subsubsection{Prediction}\vspace{-2pt}
The central two panels of the main window hold plots of the variation of the load's elevation as a function of time and the path of the elevation-velocity state vector in the capability plane through the lift. These are estimated using a computer simulation as described in \cite{Aran2011-med}, performed automatically after any of the application parameters are changed: the capability profile or the resistance settings.

\vspace{-5pt}\subsection{Discussion}\vspace{-2pt}
Lastly, we describe how the described computer tool may be used in training practice. The challenge central to the design of a continuously productive training regime is that of feedback-based adjustment of training parameters:\vspace{-5pt}
\begin{list}{\labelitemi}{\leftmargin=0.05\textwidth}
  \item[\it 1:] Training performance is with projected performance.\vspace{-8pt}
  \item[\it 2:] Limiting factors are identified.\vspace{-8pt}
  \item[\it 3:] Training parameters are suitably modified.\vspace{-8pt}
\end{list}

\vspace{-12pt}\paragraph*{Data acquisition}
It has been emphasized that one of our key aims is to develop a principle system for monitoring, evaluating and optimizing training which is inexpensive and convenient, requiring little technical proficiency from the user. Indeed the proposed methods require no more than a readily affordable camera and a PC. One of the consequences of a setup such as this is that training data can be continuously acquired, allowing for the creation of a more reliable and up-to-date model of an athlete's fitness. Specifically, video sequences (acquired by the athlete's coach, training partner or using a stationary camera set up by the athlete himself) of the athlete's training sets can be continuously fed into our capability profile estimation algorithm.

\vspace{-12pt}\paragraph*{Analysis}
The task of identifying those aspects of an athlete's fitness which are limiting performance is usually not trivial. This is because unlike the task of observing past performance, here it is necessary to be able to \emph{hypothesize} small changes to specific aspects of the athlete's fitness and furthermore \emph{predict} the nature and magnitude of performance change they would produce. The software tool described at the beginning of this section achieves precisely this in the context of resistance training. Guided by insight and experience, the coach can investigate how small changes to the athlete's current capability profile affect performance. For example, a ready estimate of the new maximal strength can be obtained. Alternatively, different training modalities can be explored. By changing the loading parameters, the practitioner can promptly see how this is reflected on the corresponding path in the capability plane i.e.\ which aspects of performance are trained the most.

\vspace{-12pt}\paragraph*{Adjustment}
The adjustment of training parameters to achieve performance improvement is intimately linked with the task of identifying those aspects of fitness which limit performance. This link is made explicit in our model and software. A productive adjustment is one which directs capability paths of training repetition sets towards capability plane regions which correspond to limiting force production conditions. This can be achieved by the practitioner though experimentation with loading parameters in the Exercise Setup Panel and observation of the effects on training performance characteristics. It is worth noting the indispensability of experience and insight, that is to say the human factor, in guiding such experimentation.

\vspace{-5pt}\section{Conclusion}\vspace{-2pt}\label{s:conc}
Motivated by the power of mathematical modelling of resistance exercise and the resulting neuromuscular adaptations, in this paper our aim was to develop a framework which would take these models from the realm of theoretical or highly specific studies and make them useful in everyday practice. Starting from raw video input, acquired using readily available, low cost equipment, the proposed framework consists of a series of steps, ending with an estimate of the parameters of the model describing a specific athlete's force production capability in a given exercise. The proposed methods were evaluated empirically using data representative of that which would be used in weight training practice. Agreement of the model's predictions with empirical performance data and relevant previous work was demonstrated. Finally, a description of a software program implementing the proposed framework was used to illustrate its possible application in practice as a tool for monitoring, evaluating, and improving training.\vspace{-6pt}

\let\oldbibliography\thebibliography
\renewcommand{\thebibliography}[1]{%
  \oldbibliography{#1}%
  \setlength{\itemsep}{0pt}%
}

\renewcommand{\baselinestretch}{0.01}
\begin{spacing}{0.0}
\fontsize{9pt}{9pt}\selectfont
\bibliographystyle{ieeetran}
\bibliography{./oa_physiology,./my_bibliography}
\end{spacing}
\end{document}